\documentclass[letterpaper, 10 pt, conference]{ieeeconf}  %

\IEEEoverridecommandlockouts                              %

\title{\LARGE \bf
Open-Fusion: Real-time Open-Vocabulary 3D Mapping and Queryable Scene Representation
}

\author{Kashu Yamazaki$^{1}$, Taisei Hanyu$^{1}$, Khoa Vo$^{1}$, Thang Pham$^{1}$, Minh Tran$^{1}$, \\ Gianfranco Doretto$^{2}$, Anh Nguyen$^{3}$, Ngan Le$^{1}$ %
\thanks{$^{1}$AICV Lab, Department of EECS, University of Arkansas, USA.
        {\tt\small {kyamazak}@uark.edu}}%
\thanks{$2$ Department of CSCE, West Virginia University, USA.}
\thanks{$3$ Department of CS, University of Liverpool, UK.}
}
\usepackage{graphicx} 
\usepackage{amssymb}
\usepackage{xcolor}
\usepackage{xspace}
\usepackage{amsmath}
\usepackage{soul}
\usepackage{hyperref}
\let\labelindent\relax
\usepackage{enumitem}
\usepackage{booktabs}
\usepackage{multicol}
\usepackage{multirow}
\usepackage[perpage]{footmisc}
\usepackage{tablefootnote}
\usepackage{pifont}%
\newcommand{\cmark}{\ding{51}}%
\newcommand{\xmark}{\ding{55}}%

\begin{document}
\setlength{\abovedisplayskip}{2pt}
\setlength{\belowdisplayskip}{1pt}
\newcommand{\model}{Open-Fusion\xspace}

\maketitle
\thispagestyle{empty}
\pagestyle{empty}

\begin{abstract}

Precise 3D environmental mapping is pivotal in robotics. Existing methods often rely on predefined concepts during training or are time-intensive when generating semantic maps. 
This paper presents Open-Fusion, a groundbreaking approach for real-time open-vocabulary 3D mapping and queryable scene representation using RGB-D data. Open-Fusion harnesses the power of a pre-trained vision-language foundation model (VLFM) for open-set semantic comprehension and employs the Truncated Signed Distance Function (TSDF) for swift 3D scene reconstruction. By leveraging the VLFM, we extract region-based embeddings and their associated confidence maps. These are then integrated with 3D knowledge from TSDF using an enhanced Hungarian-based feature-matching mechanism. Notably, Open-Fusion delivers outstanding annotation-free 3D segmentation for open-vocabulary without necessitating additional 3D training. Benchmark tests on the ScanNet dataset against leading zero-shot methods highlight Open-Fusion's superiority. Furthermore, it seamlessly combines the strengths of region-based VLFM and TSDF, facilitating real-time 3D scene comprehension that includes object concepts and open-world semantics. We encourage the readers to view the demos on our project page: \url{https://uark-aicv.github.io/OpenFusion}

\end{abstract}

\section{INTRODUCTION}

Real-time 3D scene understanding, crucial in computer vision, involves discerning object semantics, locations, and geometric attributes from RGB-D data in unstructured environments \cite{song2015sun}. Despite its diverse applications in virtual reality, robotics, and augmented reality, traditional training methods face significant challenges \cite{naseer2018indoor}. These include the need for extensive human annotations, limited closed-set semantic information, and the demand for real-time performance in applications like robotics and augmented reality.

In recent years, the convergence of language and robotics has garnered significant attention, driven by the promise it holds in enabling robots to interpret and act upon straightforward natural language commands. This benefit from the emergence of large-scale vision-language foundation models (VLFMs) such as CLIP \cite{radford2021learning}, ALIGN \cite{jia2021scaling}, BLIP \cite{li2022blip}, GLIP \cite{li2022grounded}, RegionCLIP \cite{zhong2022regionclip}, etc. Those models are learned in unsupervised manner using massive image-text pairs from the internet and  have showcased remarkable capabilities in zero-shot learning
and open-vocab reasoning. However, integrating VLFMs into robotics requires addressing scalability and real-time processing concerns. Scalability is essential to avoid exponential data growth in large environments, while real-time capability is vital for instant decision-making. Achieving these goals necessitates efficient data extraction and integration without undue delays.

Despite the impressive qualities exhibited by these VLFMs, there remains a significant untapped potential for their integration into robotic applications, particularly in the context of 3D mapping and understanding. 
The primary bottleneck in leveraging VLFMs for robotics stems from the fact that most foundation models consume images and produce only a single vector encoding of the entire image within an embedding space. This approach falls short of meeting the stringent demands of robotic perception systems, which require precise reasoning at point-level or object-level granularity across a diverse spectrum of concepts. This is crucial for tasks involving interaction with the external 3D environment, such as navigation and manipulation. Moreover, it is essential to acknowledge that applying VLFMs at the point-level can be computationally intensive and time-consuming, rendering it unsuitable for meeting the real-time demands of real-world applications. Therefore, to fully harness the potential of VLFMs in robotics, there is a pressing need to develop more efficient and effective techniques that enable these models to operate in real-time while delivering the required level of precision for tasks in complex 3D environments.

In response to the aforementioned challenges, we present \model, a queryable semantic representation rooted in VLFMs. \model facilitates real-time 3D scene reconstruction, incorporating semantics, through the use of the Truncated Signed Distance Function (TSDF). Our work demonstrates that \model excels in the efficient zero-shot reconstruction and understanding of 3D scenes, offering queryable scene representations for enhanced understanding and interaction. 
To summarize, we make the following contributions: \textbf{1) Real-time 3D Scene Reconstruction}: We extend TSDF to achieve effective real-time 3D scene reconstruction. \textbf{2) Semantic-aware Region-based Feature Matching}: We extend Hungarian matching to seamlessly match features from the VLFM into the 3D scene representation, enabling incremental semantic reconstruction.
\textbf{3) Embedding Dictionary for Efficiency}: To reduce memory consumption during scene reconstruction and facilitate open-vocab scene queries, we implement an embedding dictionary. \textbf{4) \model}: As a result, we propose \model, a real-time 3D map reconstruction and scene representation with open-vocab query capabilities. This framework promises to advance the field of real-time 3D scene understanding for robotics.

\begin{table*}[]
    \centering
    \vspace{0.5em}
    \caption{High-level comparison between our \model and existing SOTA queryable scene representations. $P$ denotes the number of points in a map, $M$ is the number of objects in the scene.}
    \vspace{-1em}
    \begin{tabular}{c|c|c|c|c|c|c|c}
    \toprule
    \textbf{Map} & \textbf{Method} & \textbf{Representation} & \textbf{Foundation Model }& \textbf{Feature Level} & \textbf{Real-time \footnotemark} & \textbf{Scene-specific} & \textbf{Sem-Query \footnotemark} \\ 
    \hline
    \multirow{3}{*}{2D} & CoW \cite{gadre2023cows} &  point & CLIP \cite{radford2021learning} + GradCAM \cite{selvaraju2022grad} & point & - & \xmark & $\mathrm{O}(P)$  \\
    & NLMap \cite{chen2023open} & point & ViLD \cite{gu2021open} + CLIP \cite{radford2021learning} & bbox & - & \xmark & $\mathrm{O}(P)$  \\
    & VLMap \cite{huang2023visual} & point & LSeg \cite{li2022language} & point & \xmark & \xmark & $\mathrm{O}(P)$  \\ \hline
    \multirow{3}{*}{3D} & CLIP-Fields \cite{shafiullah2022clip} &  NeRF & Detic  \cite{zhou2022detecting} + CLIP \cite{radford2021learning} & bbox &\xmark & \cmark & - \\
    & LERF \cite{kerr2023lerf} &  NeRF & CLIP \cite{radford2021learning} & image patch &\xmark & \cmark & - \\
    & SemAbs\cite{ha2022semantic} &  occupancy & CLIP \cite{radford2021learning} + GradCAM \cite{selvaraju2022grad} & point &\xmark & \xmark & $\mathrm{O}(P)$ \\
    & ConceptFusion \cite{jatavallabhula2023conceptfusion} & point & SAM \cite{kirillov2023segment} + CLIP \cite{radford2021learning} & bbox &\xmark & \xmark & $\mathrm{O}(P)$ \\ \cline{2-8}
    & \textbf{\model} & TSDF & SEEM \cite{zou2023segment} & region & \cmark  &\xmark & $\mathrm{O}(M)$ \\ \bottomrule 

    \end{tabular}
    \vspace{-4mm}
    \label{tab:my_label}
\end{table*}

\footnotetext[1]{Real-time denotes the real-time requirement for 3D scene reconstruction.}
\footnotetext[2]{ Query denotes the time for open-vocab semantic query.}

\section{RELATED WORKS}

\textbf{Vision-Language Foundation Models (VLFMs).} VLFMs have brought about a revolution in the field of perception by enabling open-set inference using natural language. These models, renowned for their robust generalization capabilities, owe their success to the extensive datasets and model parameters that drive them. VLFMs can be broadly categorized into three groups based on the level of resolution in vision-language alignment as follows:
\textit{(i) Image-Level Aligned Models (ILAMs)} (e.g. UniCL \cite{yang2022unified}, CLIP \cite{radford2021learning}, ALIGN \cite{jia2021scaling}, BLIP \cite{li2022blip}, and BLIPv2 \cite{li2023blip}.); \textit{(ii) Pixel-Level Aligned Models (PLAMs)} (e.g. LSeg\cite{li2022language} or MaskCLIP \cite{zhou2022extract}); \textit{(iii) Region-Level Aligned Models (RLAMs)} (e.g. GLIP \cite{li2022grounded}, GLIPv2 \cite{zhang2022glipv2}, RegionCLIP \cite{zhong2022regionclip}, ODISE \cite{xu2023open}, SEEM \cite{zou2023segment}, HIPIE \cite{wang2023hierarchical}, Semantic SAM \cite{li2023semantic}, etc). More specifically, ILAMs generate a single vector representation for the entire image by establishing connections between images and language descriptions. PLAMs and RLAMs, on the other hand, produce vector representations for  pixels and regions within an image incorporation with language descriptions, respectively.

\begin{figure*}[t]
    \centering
    \includegraphics[width=\linewidth]{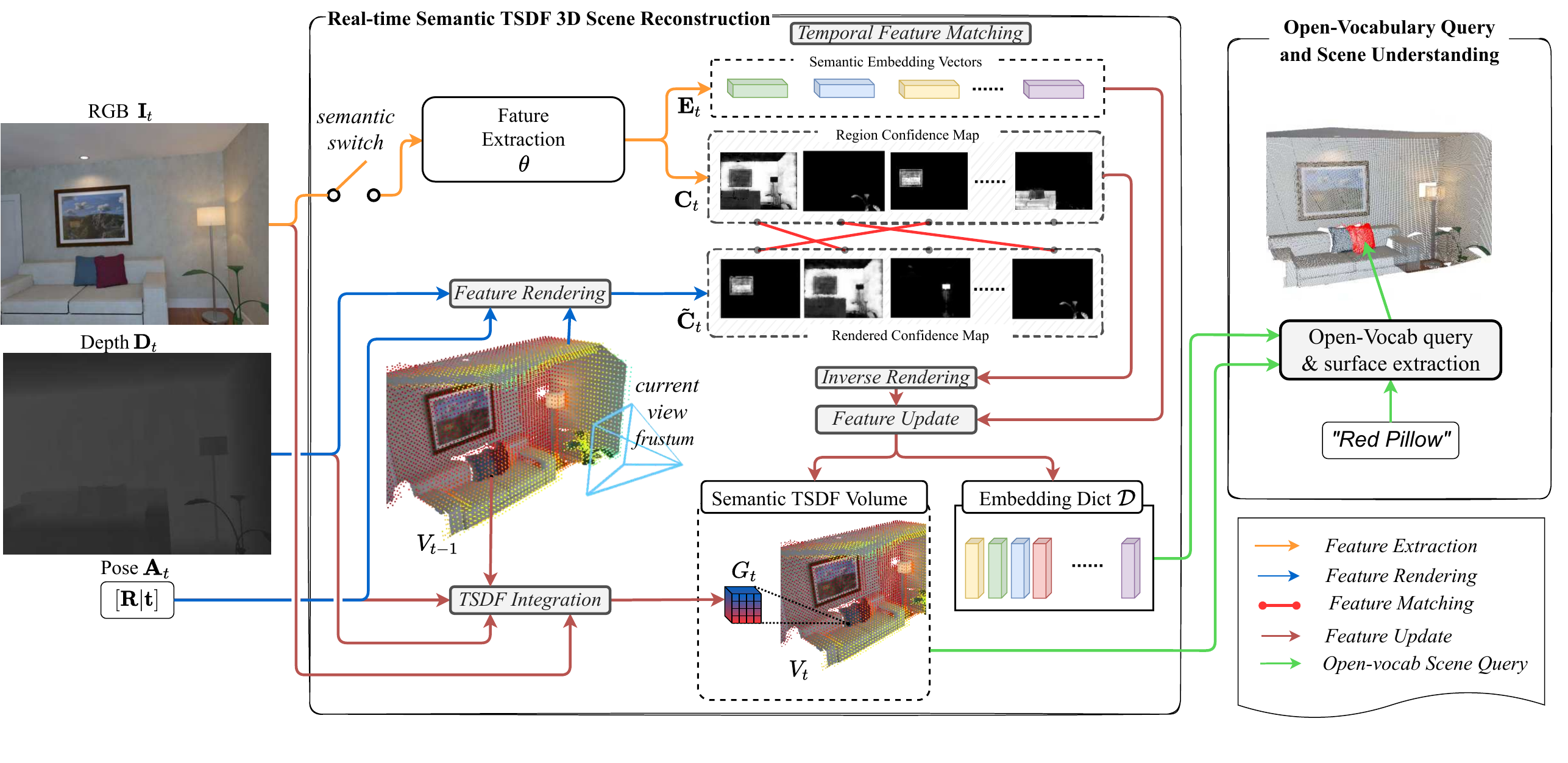}
    \vspace{-3em}
    \caption{The overall pipeline of \model, which contains two modules. \textit{Real-time Semantic TSDF 3D Scene Reconstruction Module}: This module takes in a stream of RGB-D images ($\textbf{I}_t$, $\textbf{D}_t$) and the corresponding camera pose ($\mathbf{A}_t$). It incrementally reconstructs the 3D scene, representing it as a semantic TSDF volume $V_t$ at time $t$. \textit{Open-Vocabulary Query and Scene Understanding Module}: In the second module, \model accepts open-vocab queries as inputs and provides corresponding scene segmentations in response, which can serve as an eye for language base robot commanding.
} 
\vspace{-1em}
    \label{fig:overall}
\end{figure*}

\noindent

\noindent

\noindent

While pixel-level representations offer fine semantics, they come at the cost of significant time consumption, rendering them unsuitable for real-world robotics applications where real-time processing is a necessity. Conversely, image-level representations, while efficient in terms of processing speed, provide coarser semantics, falling short in scenarios requiring fine-grained semantic comprehension, such as object-level robotics tasks like scene understanding, navigation, etc. Given this trade-off between semantic richness and computational efficiency, we have opted to leverage SEEM, a region-level VLFM with masks. SEEM strikes a balance between the demand for nuanced semantic understanding and the imperative of time-efficient processing, all while maintaining scalability.

\textbf{Queryable scene representation.} To provide a comprehensive overview of the existing scene representation, we classify prior works into three distinct categories as follows:

\noindent
\textit{2D Mapping}: CoW \cite{gadre2023cows} and NLMap \cite{chen2023open} are notable examples, harnessing the open-set features derived from CLIP to construct 2D map for exploration. CoW employs Grad-CAM \cite{selvaraju2022grad} to extract spatial knowledge from CLIP whereas NLMap integrates ViLD \cite{gu2021open} to crop objects before applying CLIP. VLMaps \cite{huang2023visual} stands out by utilizing pixel-aligned features from LSeg \cite{li2022language} to enable the creation of bird's-eye view 2D maps, specifically designed for efficient landmark querying.

\noindent
\textit{NeRF-based 3D Scene Reconstruction}: 
CLIP-Fields \cite{shafiullah2022clip} trains a NeRF-inspired implicit representation network that maps spatial coordinates $(x, y, z)$ to vectors enriched with semantic information through MLPs. Remarkably, this approach is scene-specific, with direct supervision from semantic vectors obtained from CLIP or other models like Sentence BERT \cite{reimers2019sentence}. LERF \cite{kerr2023lerf}, while also drawing inspiration from CLIP, focuses primarily on object localization. It trains a neural field through knowledge distillation from multi-scale CLIP features and DINO. However, it is worth noting that LERF may struggle with capturing precise object boundaries due to its primary emphasis on object localization.

\noindent
\textit{Non-NeRF-based 3D Scene Reconstruction}: SemAbs \cite{ha2022semantic} proposed a method to incorporate semantics from CLIP with GradCAM and the 3D completion module to produce semantic-aware occupancy. 
While it showcases promising results in 3D scene understanding, it cannot run in real-time. 
ConceptFusion \cite{jatavallabhula2023conceptfusion} introduces a unique paradigm by employing off-the-shelf foundation models to construct 3D maps with open-set features. This approach exhibits great potential for open-vocab 3D scene understanding, although it remains a point-based approach and may not meet the requirement of real-time applications.

While NeRF-based methods excel in achieving photorealistic scene reconstruction, they are not without their limitations. Notably, they require retraining for each new scene and are constrained in the volume they can render. As a result, they tend to be customized for specific scenes, which limits their applicability to real-world scenarios. Conversely, non-NeRF-based 3D scene reconstruction methods have the potential to capture more generalizable representations as they might not require retraining for each new scene. However, previous works focus on the offline generation of the queryable map mainly due to the time-consuming computational requirements posed by their point-based approach.
This drawback makes them less suitable for real-time robotics applications. 
To address this challenge 
, we introduce \model, which optimizes non-NeRF-based methods for real-time processing, resulting in a real-time open-vocab 3D scene representation.

\section{METHODOLOGY}

\subsection*{Problem Setup}
Consider a sequence of $T$ RGB-D observations obtained from an environment, which can be represented as $\{(\textbf{I}_t, \textbf{D}_t, \textbf{A}_t)\}_{t=0}^T$. Here, $\textbf{I}_t \in \mathbb{R}^{H\times W\times 3}$ represents an RGB frame, $\textbf{D}_t\in \mathbb{R}^{H\times W}$ indicates a depth frame, and $\mathbf{A}_t = [\mathbf{R}_t | \mathbf{t}_t] \in \mathbb{R}^{3 \times 4}$ denotes the associated camera pose with rotation $\mathbf{R} \in \mathbb{R}^{3 \times 3}$ and translation $\mathbf{t}\in \mathbb{R}^{3 \times 1}$. Additionally, we have the camera's intrinsic parameters represented as $\mathbf{K} \in \mathbb{R}^{3\times 3}$. Our primary objective is to construct a language-queryable 3D map denoted as $\mathcal{M}$ in real-time. In this context, we define a queryable map as a 2D/3D representation of the environment that incorporates both physical and semantic features. These features can be extracted using a query vector $\mathbf{q} \in \mathbb{R}^d$. Notably, various entities such as images, text, coordinates, etc., can be transformed into the query vector by encoding them into a shared embedding space using an encoding function.

Our proposed \model, as depicted in Fig. \ref{fig:overall}, comprises two main modules: 1) \textit{Real-time Semantic TSDF 3D Scene Reconstruction}: this module consists of two sub modules i) Feature Extraction: this module aims to extract region-based feature including confidence map and embedding map ii) Real-time Semantic 3D Scene Reconstruction: this module facilitates the integration of an incoming frame at time $t$ into the current semantic STDF volume $V_{t-1}$ while updating the embedding dictionary ($\mathcal{D}_t$). Consequently, it generates a 3D scene representation $V_T$ and an updated embedding dictionary ($\mathcal{D}_T$) after the integration of $T$ frames. The second module consists of three components of Feature Rendering by TSDF, Region-based Semantic Feature Matching, and Feature Update. 2) \textit{Open-Vocab Query and Scene Understanding}: this module is designed to localize and segment objects in the scene based on user queries and open-vocab semantics.

\subsection{Region-based Feature Extractor}
Given the RGB frame of the current view $\textbf{I}_t$ at time $t$, employ the SEEM model \cite{zou2023segment}, denoted as $\theta$, for encoding. Unlike the widely adopted CLIP model, SEEM produces \textit{region-level} aligned feature. This aims to eliminate the need for the class agnostic mask proposal generator in two-stage setup \cite{jatavallabhula2023conceptfusion} or attention-explainability model to localize the relevant regions like \cite{gadre2023cows, ha2022semantic}. Considering the real-time constraints, avoiding the inclusion of such expensive models in a sequence of function calls is of utmost importance.

For each $\textbf{I}_t$, the model $\theta$ generates region confidence maps $\mathbf{C}_t \in \mathbb{R}^{|Q| \times H/4\times W/4}$ at a quarter of the input resolution. Additionally, it produces corresponding semantic embedding vectors, denoted as $\mathbf{E}_t \in \mathbb{R}^{|Q|\times d}$, tailored for the predefined number of object queries $|Q|$, where $d$ is feature dimension. The feature extraction at time $t$ can be formulated as $\mathbf{C}_t, \mathbf{E}_t = \theta(\mathbf{I}_t)$. In practice, the region-based feature extraction process is specifically for semantic-related tasks and may pose a bottleneck due to SEEM's time consumption at 4.5 FPS. If a task doesn't require semantics, this process can be skipped. Additionally, given the substantial overlap between two consecutive frames, it's possible to skip some frames. To enhance the flexibility and efficiency of our OpenFusion, we have implemented a semantic switch, as depicted in Fig. 1.

\subsection{Real-time 3D Scene Reconstruction with Semantics}
\label{sec:3D_scene}
Every time-frame, we incorporate the incoming observation $(\textbf{I}_t, \textbf{D}_t, \mathbf{A}_t)$ into an implicit surface using the Truncated Signed Distance Function (TSDF). Specifically, we integrate $(\textbf{I}_t, \textbf{D}_t, \mathbf{A}_t)$ into the TSDF volume $V_{t-1}$ to create the TSDF volume at time $t$, denoted as $V_t$. It is important to emphasize that the TSDF volume $V_{t}$ comprises a set of $M$ volumetric blocks, represented as $V_t = \{G_i\}_{i=1}^M$.
The TSDF is an extension of the Signed Distance Function (SDF) $\phi$, which is a function that provides the shortest distance to any surface for every 3D point. The sign indicates whether the point is located in front of or behind the surface. In the context of scene reconstruction, the points of interest typically reside on the boundary $\delta \Omega$. For a distance function $d$ and a point $p \in \mathbb{R}^3$, the SDF $\phi: \mathbb{R}^3 \rightarrow \mathbb{R}$ defines the signed distance to the surface as follows:
\begin{equation}
    \phi(p) = \left\{\begin{matrix}
    -d(p, \delta\Omega) & \text{if }p \in \Omega\\ 
 d(p, \delta\Omega) & \text{if } p \in \Omega^c
\end{matrix}\right.
\end{equation}

This means that points located inside the surface have negative values, while the surface itself lies precisely at the zero crossing point between positive and negative values. The TSDF truncates all values above with a specified threshold $\tau$, with $\tau$ chosen as four times the voxel size.

As the reconstruction of the 3D scene essentially represents a local 2D surface—a 2D manifold embedded in 3D space—we can efficiently embed the 3D scene using globally sparse but locally dense voxel blocks. These voxel blocks exhibit a distinctive characteristic where they are globally present only near the surface of interest (while other parts remain void). Within each block, we maintain a dense voxel grid typically sized at $r \times r \times r$. Following the approach in \cite{dong2019gpu}, we construct \emph{semantic TSDF volume} as a set of globally sparse volumetric blocks $V_t = \{G_i\}_{i=1}^M$, each containing a locally dense voxel grid $G_i = \{p_j\}_{j=1}^{r^3}$ and  the information in $p_j$ include $p_j = \{(RGB_j, w_j, \phi_j, k_j, c_j)\}$. These grids store various attributes of a voxel $p_j$, including color $RGB$, weight $w$ for TSDF updates, TSDF values $\phi$, embedding keys $k$, and confidence scores $c$.

Notably, unlike previous approaches like \cite{jatavallabhula2023conceptfusion} that store the semantic embedding for each point, we opt to store only the keys for embedding and the associated confidence scores for each pixel. The actual embedding information is maintained separately within the dedicated embedding dictionary $\mathcal{D}$. Given our utilization of region-based embedding for the scene, it's important to highlight that the number of embeddings required for the entire scene is significantly smaller compared to point-based counterparts. 
In addition to the surface and color data, we also incorporate semantics into the TSDF volume. However, to optimize computation and memory usage in subsequent modules, we limit the storage of semantics to points near the surface. These points are strategically sampled based on the TSDF values, resulting in a more efficient representation.

As a result, to integrate $(\textbf{I}_t, \textbf{D}_t, \mathbf{A}_t)$ at time $t$ into semantic TSDF volume $V_{t-1}$ at time $t-1$, consisting of $M$ volumetric blocks $\{G_i\}_{i=1}^M$, we perform the following steps. Step (i) - \textit{Feature Rendering}: This initial step involves generating a rendered confidence map $\Tilde{\mathbf{C}}_t$ and a rendered embedding $\Tilde{\mathbf{E}}_t$ from the existing TSDF volume $V_{t-1}$. Step (ii) - \textit{Region-based Temporal Feature Matching}: In this phase, the goal is to establish correspondences between the semantic embedding vector $\mathbf{E}_t$ from frame $t$ and the rendered embedding $\Tilde{\mathbf{E}}_t$. Step (iii) - \textit{Feature Update}: This step focuses on updating the TSDF volume $V_{t-1}$ at time $t$ to create $V_t$ and concurrently updating the embedding dictionary ($\mathcal{D}_t$).

\subsubsection{Feature Rendering by TSDF}

We render confidence map $\Tilde{\mathbf{C}}_t$ with its corresponding embeding map $\Tilde{\mathbf{E}}_t$ from the TSDF volume with the current camera pose $\mathbf{R}_t|\mathbf{t}_t$ and depth image $\textbf{D}_t$ at time $t$.
Given the semantic TSDF volume $V_{t-1}$ accumulated from time $0$ to $t-1$ and the current observation $(\textbf{I}_t, \textbf{D}_t, \mathbf{A}_t)$, our integration process involves several key steps: i) Conversion of depth image $\textbf{D}_t$: Initially, we convert the 2D depth image $\textbf{D}_t$ into a 3D representation using Eq.\ref{eq:2D-3D}.

\begin{equation}
    \left[\begin{array}{l}
x \\
y \\
z
\end{array}\right]=\mathbf{R}_t^{-1} \left( \mathbf{D}^{i, j}_t \mathbf{K}^{-1} \left[\begin{array}{l}
i \\
j \\
1
\end{array}\right] \right) - \mathbf{t}_t \; ,
\label{eq:2D-3D}
\end{equation}

\noindent
where $\mathbf{R}_t$ and $t_t$ are the rotation and translation component of camera pose $\mathbf{A}_t$, and $\mathbf{K}$ represents intrinsic parameters. ii) Identifying relevant blocks: Next, we identify the set of volumetric blocks $\mathcal{G}_{active}$ that contain points unprojected from the current depth image. We determine these active blocks within the current viewing frustum by examining whether the 3D coordinates $(x, y, z)$, obtained by casting a ray from the camera's origin through the pixel $(i, j)$ of the depth image $\textbf{D}_t$, fall within the boundaries of these blocks or not. 
iii) Projection of semantic information: Subsequently, we project the voxels within the active blocks ${G}_j$ that possess semantic keys and confidence scores onto the image plane, as defined by Eq.\ref{eq:render}.

\begin{equation}
    \left[\begin{array}{l}
u \\
v \\
\hat{d}
\end{array}\right] = \hat{\mathbf{K}} \left(\mathbf{R}_t \left[\begin{array}{l}
x \\
y \\
z
\end{array}\right] +\mathbf{t}_t\right) \; ,
\label{eq:render}
\end{equation}

\noindent
where $\hat{\mathbf{K}}$ represents the intrinsic parameters with the model $\theta$'s output reslution and the coordinate of the valid voxel $(x,y,z)$ are mapped to the pixel location $(u/\hat{d}, v/\hat{d})$ subjected to $\left( \hat{d} > 0 \right) \land \left( 0 \leq u/\hat{d} < W/4 \right) \land \left( 0 \leq v/d' < H/4 \right)$. 

This projection is a crucial step in incorporating semantic information into the current frame's representation. Building upon the rendering operation described above, we generate confidence maps $\Tilde{\mathbf{C}}_t\in \mathbb{R}^{m \times H/4\times W/4}$ within the current field of view (FoV).

\subsubsection{Region-based Temporal Feature Matching}

This step aims to find fusion candidates by matching pairs between the confidence map $\mathbf{C}_t$ and the rendered confidence map $\Tilde{\mathbf{C}}_t$, which casts the knowledge of objects accumulated until $t-1$ in the semantic TSDF volume from the current FoV. We formulate this feature matching as a 2D rectangular assignment problem, with the goal of identifying the assignment $\mathcal{S}^*$ that maximizes the soft-IoU \cite{huang2019batching} between $\mathbf{C}_t$ and $\Tilde{\mathbf{C}}_t$.

\begin{equation}
    \mathcal{S}^*=\underset{\mathcal{S}}{\arg \max } \sum_{i=1}^n \sum_{j=1}^m\mathcal{L}_{match} \langle \mathbf{C}_t, \Tilde{\mathbf{C}}_t\rangle_{i,j}\sigma_{i,j} \; ,
\end{equation}

\noindent Here, $n$ represents the number of semantic regions in the current frame, and $m$ is the number of rendered regions within the current FoV. $\mathcal{L}_{match}$ calculates the soft-IoU of $\mathbf{C}_t$ and $\Tilde{\mathbf{C}}_t$. The matrix $\mathcal{S}$ represents a set of all $\sigma_{i,j}$ values, subject to the constraints $\sum_{j=1}^m \sigma_{i,j} = 1 ; \forall i$, $\sum_{i=1}^n \sigma_{i,j} \leq 1 ; \forall j$, and $\sigma_{i,j} \in \{0,1\}$. If $\sigma_{i,j} = 1$, it signifies that the prediction in row $i$ is assigned to the rendered embedding in column $j$. To solve this problem, we employ a modified Jonker-Volgenant algorithm \cite{crouse2016implementing} (extension version of Hungarian). 
We discard the match if the soft-IoU score is below $0.10$. This operation helps us to avoid fusing poor quality masks of the same object due to occlusion or blur.

\subsubsection{Inverse Rendering and Feature Update}
In this step, information, i.e., $(\textbf{I}_t, \textbf{D}_t, \mathbf{C}_t, \mathbf{E}_t)$, we obtain from the current time frame is integrated into the semantic TSDF volume $V_{t-1}$ to create $V_{t}$. First, each voxel $p_j$ within the active volumetric blocks $\mathcal{G}_{active}$ undergoes the standard TSDF integration process \cite{dong2019gpu}, where the stored color $RGB_j$ and TSDF values $\phi_j$ are updated using weighted average. Using Eq. \ref{eq:render} with the actual camera intrinsic $\mathbf{K}$, we can obtain $(u,v,d)$ and the update is summarized as:
\begin{equation}
RGB_j \leftarrow \frac{w_j\cdot RGB_j + \textbf{I}_t^{u/d, v/d}}{w_j + 1}
\end{equation}
\begin{equation}
\phi_j \leftarrow \frac{w_j\cdot \phi_j + \Psi(\textbf{D}_t^{u/d, v/d} - d, \tau)}{w_j + 1}
\end{equation}
\begin{equation}
w_j \leftarrow w_j + 1
\end{equation}
where $\Psi(\cdot)$ is the truncation operation that is applied to SDF to obtain TSDF (Section \ref{sec:3D_scene}).  

The dictionary $\mathcal{D}$ and the confidence score $c_j$ and the associated key $k_j$ will be updated according to the matching $\mathcal{S}^*$. If the new region is matched to the existing one, only the confidence map is updated with  weighted average while unmatched candidates also update the dictionary as a new region. The confidence maps $\mathbf{C}_t$ are inversely rendered by applying Eq.\ref{eq:render}.

\subsection{Querying Semantics from the 3D Map}

At any time $t$, we can extract the corresponding point cloud or mesh from the semantic TSDF volume $V_t$ by querying it with a vector $\mathbf{q}$. Our querying method involves a similarity calculation between the query and the semantic embeddings stored in the dictionary $\mathcal{D}_t$. This approach is significantly faster and more memory-efficient than previous methods that store embeddings for individual points. Specifically, we calculate the cosine similarity $\cos \langle \mathbf{E}, \mathbf{q}\rangle$ between the semantic embeddings $\mathbf{E} \in \mathbb{R}^{R \times d}$ in the dictionary $\mathcal{D}_t$ and the query vector $\mathbf{q} \in \mathbb{R}^{d}$, which is obtained using a modality-specific encoder trained in a shared embedding space with the semantic vectors $\mathbf{E}$, and select the most relevant region as the object proposal. 
After the query, Marching Cubes is applied to extract surfaces or point clouds from the semantic TSDF volumes to indicate the queried region. For a resource constraint environment, one can simply use the semantic TSDF voxel coordinates as the approximation of the region.

\begin{table}
    \centering
    \setlength{\tabcolsep}{2pt}
    \renewcommand{\arraystretch}{1.0}
    \vspace{0.5em}
    \caption{Quantitative comparison of \textbf{open-set semantic segmentation} and \textbf{3D scene representation time} between \model and existing methods on the ScanNet dataset.}
    \vspace{-1em}
    \resizebox{\columnwidth}{!}{%
    \begin{tabular}{l|l|cc|cc}
    \toprule
    ~ & \multirow{2}{*}{\textbf{Method}} & \multicolumn{2}{c|}{\textbf{Time (FPS)$\uparrow$}}&  \multicolumn{2}{c}{\textbf{Accuracy$\uparrow$}} \\
    & & \textbf{3D-Rec.}\footnotemark & \textbf{Sem-3D-Rec}\footnotemark & \textbf{mAcc} & \textbf{f-mIoU}
    \\ \hline
        \multirow{5}{*}{\rotatebox{90}{Priv.} \footnotemark} & LSeg & - & -& 0.70 & 0.63 \\ 
         & OpenSeg & - & -& 0.63 & 0.62 \\ 
        & CLIPSeg (rd64-uni) & - & -& 0.41 & 0.34 \\ 
        & CLIPSeg (rd16-uni) & - & -& 0.41 & 0.36 \\ 
        \hline
         \multirow{3}{*}{\rotatebox{90}{ZS.} \footnotemark} & MaskCLIP & - & -& 0.24 & 0.28 \\ 
        & ConceptFusion & 1.5 & 0.15 & \textbf{0.63} & 0.58 \\ 
        & \textbf{\model} & \textbf{50} & \textbf{4.5} & 0.62 & \textbf{0.59} \\ \bottomrule
    \end{tabular}}
    \label{tab:scan_net}
\end{table}
\footnotetext[1]{3D-Rec.: 3D scene reconstruction only.}
\footnotetext[2]{Sem-3D-Rec: 3D scene reconstruction with semantics} 
\footnotetext[3]{Priv.: off-the-shelf VLFMs are finetuned specifically for semantic segmentation.}
\footnotetext[4]{ZS: zero-shot approaches.}

\section{EXPERIMENTS}

In this section, we conduct a comprehensive evaluation of \model's performance through both quantitative and qualitative assessments on the ScanNet \cite{dai2017scannet} and Replica \cite{straub2019replica} datasets, specifically focusing on open-set semantic segmentation tasks. 
Following the previous work \cite{jatavallabhula2023conceptfusion}, we will focus our quantitative results and comparisons exclusively on the ScanNet dataset.
We will provide qualitative results and comparisons for the Replica dataset. Furthermore, we showcase the real-world applicability of \model by seamlessly integrating it into the Kobuki platform, enabling real-time 3D scene representation.

\subsection{Quantitative Benchmarks} 

Our quantitative experimental benchmarks are conducted on the ScanNet dataset, a comprehensive RGB-D video dataset comprising more than 2.5 million views across over 1,500 scans. This dataset includes annotations for 3D camera poses, surface reconstructions, and instance-level semantic segmentations. Following the methodology of ConceptFusion \cite{jatavallabhula2023conceptfusion}, we select 20 indoor (room-scale) scenes for rigorous evaluation and testing of our research. For each selected scene, we utilize the semantic categories provided in the scene annotations as text-prompted queries for performing open-set segmentation tasks.

Consistent with the evaluation methodology introduced in ConceptFusion \cite{jatavallabhula2023conceptfusion}, we assess both performance and time efficiency of \model in the context of open-set semantic 3D scene understanding. Our evaluation encompasses a dual focus: performance and time efficiency. To assess accuracy, we employ the mean accuracy (mAcc) and frequency mean Intersection over Union (f-mIoU) metrics. In addition, we measure time consumption for 3D scene representation in frames per second (FPS).
Table \ref{tab:scan_net} offers a comprehensive comparative analysis between \model against existing SOTA methods in terms of mAcc, f-mIoU, and FPS. Thanks to the utilization of region-based embedding and TSDF, \model achieves nearly real-time performance at 4.5 FPS, which is 30 times faster than the runner-up ConceptFusion. While excelling in time efficiency, \model maintains competitive performance levels with the existing SOTA ConceptFusion in terms of both mAcc and f-mIoU metrics. 
This experiment underscores the efficiency and effectiveness of \model in the realm of open-set semantic 3D scene understanding. \model represents a significant advancement, establishing itself as the new SOTA in terms of both performance and efficiency.

\subsection{Qualitative Results}
\begin{figure}
    \centering
    \vspace{1mm}
    \includegraphics[width=\linewidth]{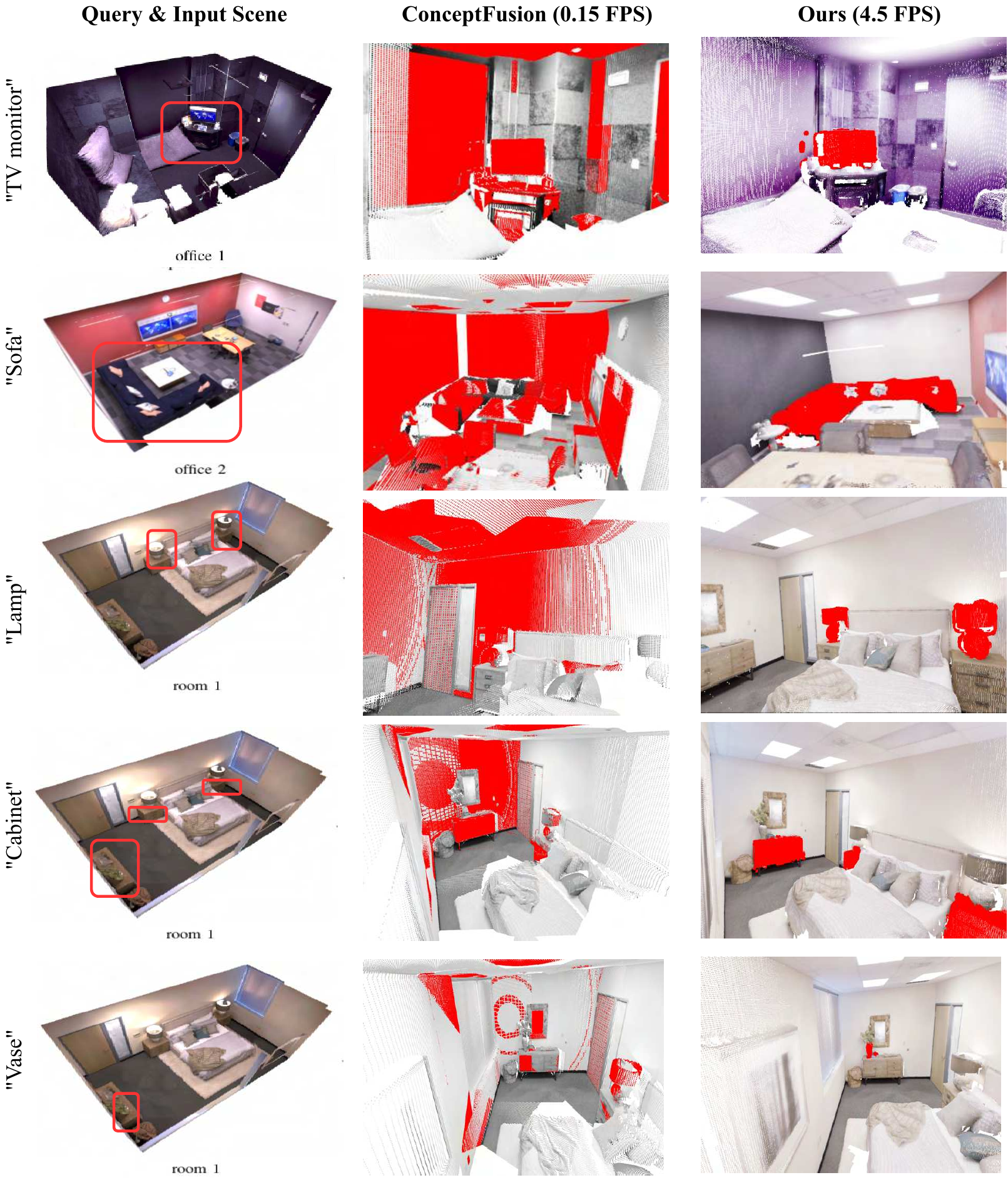}
    \caption{Qualitative comparison of 3D object query results on Replica dataset. While ConceptFusion failed to pinpoint the object location, \model can estimate more precise location from language queries.}
    \vspace{-1em}
    \label{fig:Replica}
    \vspace{-3mm}
\end{figure}

We conducted a qualitative evaluation of \model on the Replica dataset \cite{straub2019replica}, as illustrated in Fig. \ref{fig:Replica}. In this experiment, we demonstrated the semantic segmentation performance with queries involving various object sizes, from small objects like vases and lamps to larger ones like sofas and cabinets. Our \model not only achieves significantly faster processing times (30x faster) but also delivers more accurate queryable semantic segmentation results.

\subsection{Real-World Experiment}
\begin{figure}[!t]
    \centering
    \includegraphics[width=\linewidth]{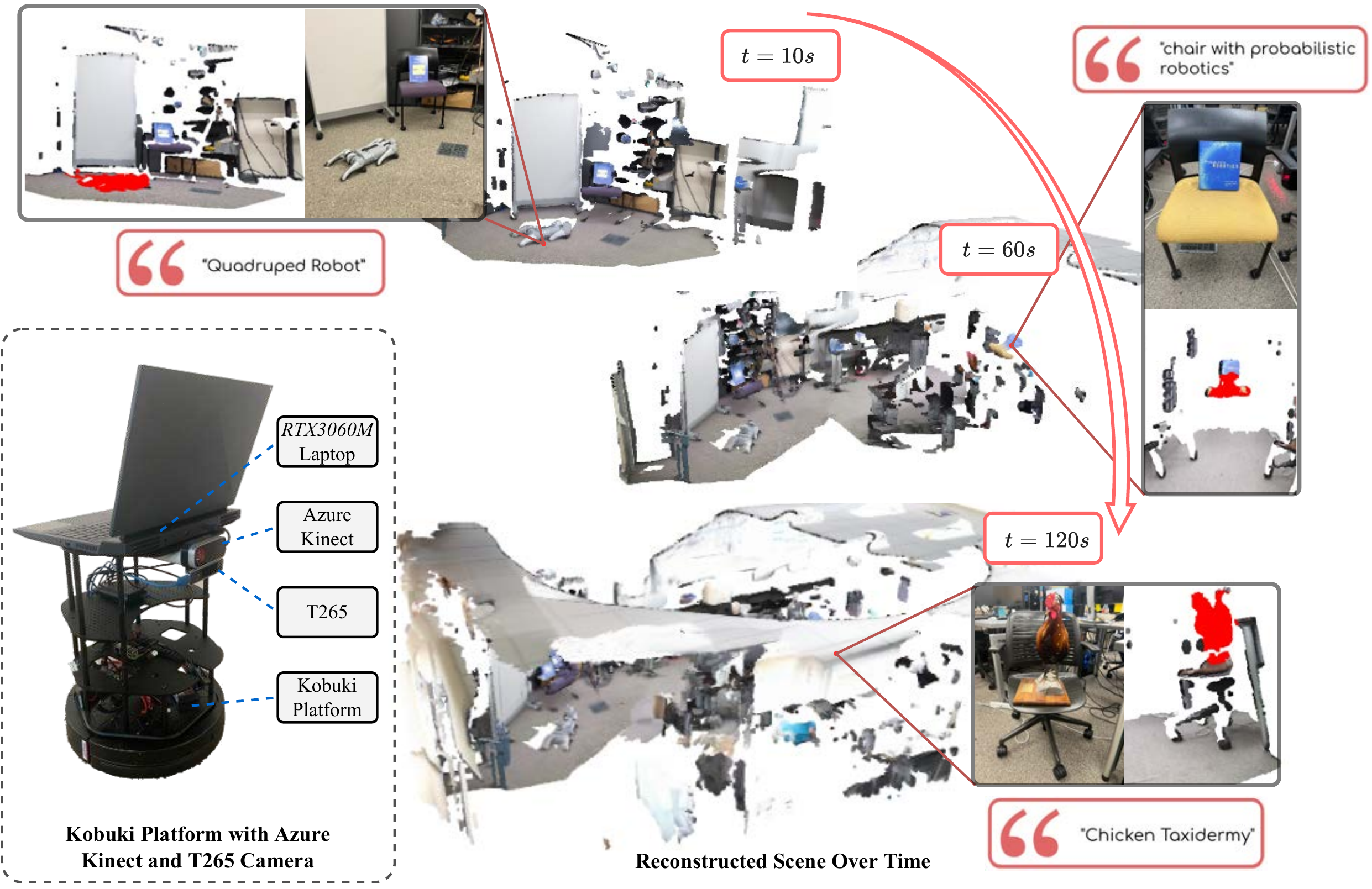}
    \vspace{-2em}
    \caption{The Kobuki platform is equipped with an Azure Kinect Camera and an Intel T265 Camera to demonstrate real-time mapping in a real-world environment. This system enables interaction with the world through natural language queries. The system is able to highlight the novel objects like the "quadruped robot" or "chicken taxidermy".}
    \vspace{-1em}
    \label{fig:Kobuki_res}
\end{figure}

In this section, we present a real-world demonstration of real-time queryable scene reconstruction. Our experiment was conducted using the Kobuki platform, equipped with an RGB-D camera setup. Specifically, we utilized the \textit{Azure Kinect Camera} to capture RGB-D images at a downsampled resolution of $360 \times 630$, along with the \textit{Intel T265 Camera} for capturing corresponding camera poses. To ensure accurate alignment between camera poses and image streams, we synchronized them based on timestamps and filtered out any images without a matching pose recorded within a $10$ ms timeframe. Fig. \ref{fig:Kobuki_res} provides a visual representation of our real-world experimental setup using the Kobuki platform.

As it is difficult to obtain the ground truth semantic mask for real environment, we visually compare the suggested region by the model with the known environment setup. Fig. \ref{fig:Kobuki_res} displays the 3D map reconstruction generated at 50 FPS and semantics updated at 4.5 FPS running on two threads by the Kobuki platform. In this demonstration, we emphasize specific objects within the scene and their corresponding 3D representations.

\noindent
\section{CONCLUSION \& DISCUSSION}

In this paper, we have introduced \model, an efficient approach for real-time open-vocabulary 3D mapping and queryable scene representation from RGB-D data. \model leverages the VLFM, SEEM, to extract region-based embeddings and employs TSDF, along with an extended version of Hungarian matching, for 3D semantic representation. We conducted both qualitative and quantitative benchmarks to assess our performance. In a qualitative evaluation, we compared \model with ConceptFusion using the Replica dataset, demonstrating superior object segmentation results and real-time efficiency. In a quantitative assessment, we compared \model with SOTA methods using the ScanNet dataset, achieving competitive results in terms of mean accuracy (mAcc) and surface mean Intersection over Union (f-mIoU) while \model is 30x faster than ConceptFusion. Additionally, we conducted a real-world experiment with the Kobuki platform, highlighting \model's capability in practical applications.

It is worth noting that while \model excels in geometric precision, it may not capture the full range of photometric nuances due to the constraints of TSDF representations. This limitation will be potential future improvement.

\newpage
\bibliographystyle{ieeetr}
\bibliography{main.bib}

\begin{thebibliography}{10}

\bibitem{song2015sun}
S.~Song, S.~P. Lichtenberg, and J.~Xiao, ``Sun rgb-d: A rgb-d scene
  understanding benchmark suite,'' in {\em Proceedings of the IEEE conference
  on computer vision and pattern recognition}, pp.~567--576, 2015.

\bibitem{naseer2018indoor}
M.~Naseer, S.~Khan, and F.~Porikli, ``Indoor scene understanding in 2.5/3d for
  autonomous agents: A survey,'' {\em IEEE access}, vol.~7, pp.~1859--1887,
  2018.

\bibitem{radford2021learning}
A.~Radford, J.~W. Kim, C.~Hallacy, A.~Ramesh, G.~Goh, S.~Agarwal, G.~Sastry,
  A.~Askell, P.~Mishkin, J.~Clark, {\em et~al.}, ``Learning transferable visual
  models from natural language supervision,'' in {\em International conference
  on machine learning}, pp.~8748--8763, PMLR, 2021.

\bibitem{jia2021scaling}
C.~Jia, Y.~Yang, Y.~Xia, {\em et~al.}, ``Scaling up visual and vision-language
  representation learning with noisy text supervision,'' in {\em ICLR},
  pp.~4904--4916, PMLR, 2021.

\bibitem{li2022blip}
J.~Li, D.~Li, C.~Xiong, and S.~Hoi, ``Blip: Bootstrapping language-image
  pre-training for unified vision-language understanding and generation,'' in
  {\em International Conference on Machine Learning}, pp.~12888--12900, PMLR,
  2022.

\bibitem{li2022grounded}
L.~H. Li, P.~Zhang, H.~Zhang, J.~Yang, C.~Li, Y.~Zhong, L.~Wang, L.~Yuan,
  L.~Zhang, J.-N. Hwang, {\em et~al.}, ``Grounded language-image
  pre-training,'' in {\em Proceedings of the IEEE/CVF Conference on Computer
  Vision and Pattern Recognition}, pp.~10965--10975, 2022.

\bibitem{zhong2022regionclip}
Y.~Zhong, J.~Yang, {\em et~al.}, ``Regionclip: Region-based language-image
  pretraining,'' in {\em CVPR}, pp.~16793--16803, 2022.

\bibitem{gadre2023cows}
S.~Y. Gadre, M.~Wortsman, G.~Ilharco, L.~Schmidt, and S.~Song, ``Cows on
  pasture: Baselines and benchmarks for language-driven zero-shot object
  navigation,'' in {\em Proceedings of the IEEE/CVF Conference on Computer
  Vision and Pattern Recognition}, pp.~23171--23181, 2023.

\bibitem{selvaraju2022grad}
R.~Selvaraju, M.~Cogswell, A.~Das, R.~Vedantam, D.~Parikh, and D.~Batra,
  ``Grad-cam: Visual explanations from deep networks via gradient-based
  localization. arxiv 2016,'' {\em arXiv preprint arXiv:1610.02391}, 2022.

\bibitem{chen2023open}
B.~Chen, F.~Xia, B.~Ichter, K.~Rao, K.~Gopalakrishnan, M.~S. Ryoo, A.~Stone,
  and D.~Kappler, ``Open-vocabulary queryable scene representations for real
  world planning,'' in {\em 2023 IEEE International Conference on Robotics and
  Automation (ICRA)}, pp.~11509--11522, IEEE, 2023.

\bibitem{gu2021open}
X.~Gu, T.-Y. Lin, W.~Kuo, and Y.~Cui, ``Open-vocabulary object detection via
  vision and language knowledge distillation,'' {\em arXiv preprint
  arXiv:2104.13921}, 2021.

\bibitem{huang2023visual}
C.~Huang, O.~Mees, A.~Zeng, and W.~Burgard, ``Visual language maps for robot
  navigation,'' in {\em 2023 IEEE International Conference on Robotics and
  Automation (ICRA)}, pp.~10608--10615, IEEE, 2023.

\bibitem{li2022language}
B.~{Li}, K.~Q. {Weinberger}, S.~{Belongie}, V.~{Koltun}, and R.~{Ranftl},
  ``{Language-driven Semantic Segmentation},'' {\em arXiv e-prints},
  p.~arXiv:2201.03546, Jan. 2022.

\bibitem{shafiullah2022clip}
N.~M.~M. Shafiullah, C.~Paxton, L.~Pinto, S.~Chintala, and A.~Szlam,
  ``Clip-fields: Weakly supervised semantic fields for robotic memory,'' {\em
  arXiv preprint arXiv:2210.05663}, 2022.

\bibitem{zhou2022detecting}
X.~Zhou, R.~Girdhar, A.~Joulin, P.~Kr{\"a}henb{\"u}hl, and I.~Misra,
  ``Detecting twenty-thousand classes using image-level supervision,'' in {\em
  European Conference on Computer Vision}, pp.~350--368, Springer, 2022.

\bibitem{kerr2023lerf}
J.~Kerr, C.~M. Kim, K.~Goldberg, A.~Kanazawa, and M.~Tancik, ``Lerf: Language
  embedded radiance fields,'' {\em arXiv preprint arXiv:2303.09553}, 2023.

\bibitem{ha2022semantic}
H.~Ha and S.~Song, ``Semantic abstraction: Open-world 3d scene understanding
  from 2d vision-language models,'' in {\em 6th Annual Conference on Robot
  Learning}, 2022.

\bibitem{jatavallabhula2023conceptfusion}
K.~M. Jatavallabhula, A.~Kuwajerwala, Q.~Gu, M.~Omama, T.~Chen, S.~Li, G.~Iyer,
  S.~Saryazdi, N.~Keetha, A.~Tewari, {\em et~al.}, ``Conceptfusion: Open-set
  multimodal 3d mapping,'' {\em arXiv preprint arXiv:2302.07241}, 2023.

\bibitem{kirillov2023segment}
A.~Kirillov, E.~Mintun, N.~Ravi, H.~Mao, C.~Rolland, L.~Gustafson, T.~Xiao,
  S.~Whitehead, A.~C. Berg, W.-Y. Lo, {\em et~al.}, ``Segment anything,'' {\em
  arXiv preprint arXiv:2304.02643}, 2023.

\bibitem{zou2023segment}
X.~Zou, J.~Yang, H.~Zhang, F.~Li, L.~Li, J.~Gao, and Y.~J. Lee, ``Segment
  everything everywhere all at once,'' {\em arXiv preprint arXiv:2304.06718},
  2023.

\bibitem{yang2022unified}
J.~Yang, C.~Li, {\em et~al.}, ``Unified contrastive learning in
  image-text-label space,'' in {\em CVPR}, pp.~19163--19173, 2022.

\bibitem{li2023blip}
J.~Li, D.~Li, S.~Savarese, and S.~Hoi, ``Blip-2: Bootstrapping language-image
  pre-training with frozen image encoders and large language models,'' {\em
  arXiv preprint arXiv:2301.12597}, 2023.

\bibitem{zhou2022extract}
C.~Zhou, C.~C. Loy, and B.~Dai, ``Extract free dense labels from clip,'' in
  {\em European Conference on Computer Vision}, pp.~696--712, Springer, 2022.

\bibitem{zhang2022glipv2}
H.~Zhang, P.~Zhang, {\em et~al.}, ``Glipv2: Unifying localization and
  vision-language understanding,'' {\em NIPS}, 2022.

\bibitem{xu2023open}
J.~Xu, S.~Liu, A.~Vahdat, W.~Byeon, X.~Wang, and S.~De~Mello, ``Open-vocabulary
  panoptic segmentation with text-to-image diffusion models,'' in {\em
  Proceedings of the IEEE/CVF Conference on Computer Vision and Pattern
  Recognition}, pp.~2955--2966, 2023.

\bibitem{wang2023hierarchical}
X.~Wang, S.~Li, K.~Kallidromitis, Y.~Kato, K.~Kozuka, and T.~Darrell,
  ``Hierarchical open-vocabulary universal image segmentation,'' {\em arXiv
  preprint arXiv:2307.00764}, 2023.

\bibitem{li2023semantic}
F.~Li, H.~Zhang, P.~Sun, X.~Zou, S.~Liu, J.~Yang, C.~Li, L.~Zhang, and J.~Gao,
  ``Semantic-sam: Segment and recognize anything at any granularity,'' {\em
  arXiv preprint arXiv:2307.04767}, 2023.

\bibitem{reimers2019sentence}
N.~Reimers and I.~Gurevych, ``Sentence-bert: Sentence embeddings using siamese
  bert-networks,'' {\em arXiv preprint arXiv:1908.10084}, 2019.

\bibitem{dong2019gpu}
W.~Dong, J.~Park, Y.~Yang, and M.~Kaess, ``Gpu accelerated robust scene
  reconstruction,'' in {\em 2019 IEEE/RSJ International Conference on
  Intelligent Robots and Systems (IROS)}, pp.~7863--7870, IEEE, 2019.

\bibitem{huang2019batching}
Y.~Huang, Z.~Tang, D.~Chen, K.~Su, and C.~Chen, ``Batching soft iou for
  training semantic segmentation networks,'' {\em IEEE Signal Processing
  Letters}, vol.~27, pp.~66--70, 2019.

\bibitem{crouse2016implementing}
D.~F. Crouse, ``On implementing 2d rectangular assignment algorithms,'' {\em
  IEEE Transactions on Aerospace and Electronic Systems}, vol.~52, no.~4,
  pp.~1679--1696, 2016.

\bibitem{dai2017scannet}
A.~Dai, A.~X. Chang, M.~Savva, M.~Halber, T.~Funkhouser, and M.~Nie{\ss}ner,
  ``Scannet: Richly-annotated 3d reconstructions of indoor scenes,'' in {\em
  Proceedings of the IEEE conference on computer vision and pattern
  recognition}, pp.~5828--5839, 2017.

\bibitem{straub2019replica}
J.~Straub, T.~Whelan, L.~Ma, Y.~Chen, E.~Wijmans, S.~Green, J.~J. Engel,
  R.~Mur-Artal, C.~Ren, S.~Verma, {\em et~al.}, ``The replica dataset: A
  digital replica of indoor spaces,'' {\em arXiv preprint arXiv:1906.05797},
  2019.

\end{thebibliography}

\end{document}